\documentclass[letterpaper, 10 pt, conference]{ieeeconf}

\IEEEoverridecommandlockouts    % If you want to use the thanks command

\usepackage{cite} 
\usepackage{graphicx}
\usepackage{amsmath,amssymb}
\usepackage{booktabs}
\usepackage{multirow}
\usepackage{algorithm}
\usepackage{algorithmic}
\usepackage{url}
\usepackage{xcolor}
\usepackage{tikz}
\usepackage{comment}
\usepackage{censor}
\usetikzlibrary{arrows.meta,positioning}
\usepackage[hidelinks]{hyperref}

\StopCensoring

\title{\LARGE \bf Room-Mediated Co-occurrence for Zero-Shot Object-Centric Semantic Navigation via Frontier Scoring}

\author{\censor{Adam Scicluna}, \censor{Gavin Paul} and \censor{Alen Alempijevic}%
\thanks{All authors are with the \xblackout{Robotics Institute, Faculty of Engineering and Information Technology, University of Technology Sydney (UTS), Sydney, Australia}. Corresponding author: {\tt\footnotesize \{\censor{adam.scicluna@uts.edu.au}\}}}
\thanks{\censor{Adam Scicluna} is supported by \xblackout{Australian Government Research Training Program Scholarships}.}
\thanks{© 2026 IEEE.  Personal use of this material is permitted.  Permission from IEEE must be obtained for all other uses, in any current or future media, including reprinting/republishing this material for advertising or promotional purposes, creating new collective works, for resale or redistribution to servers or lists, or reuse of any copyrighted component of this work in other works.}
}

\begin{document}
\maketitle

\begin{abstract}
Zero-shot ObjectNav methods increasingly use vision-language priors, but direct object-object similarity in the latent space is often a weak proxy for spatial co-occurrence. We present an analytical, training-free semantic navigation pipeline that mediates object relationships through a compact room lexicon. Each object label is mapped to a CLIP-derived \emph{Room Probability Vector} (RPV), and object-target co-occurrence is computed from RPV distribution overlap. 
These scores are projected onto a value map using geodesic flood-fill propagation (Fast Marching Method), with adaptive signal decay, and are used to rank frontiers by semantic score for navigation. Together, these components form an integrated, training‑free, object‑centric pipeline for open‑vocabulary zero‑shot navigation. Results show that our object-centric approach improves Success Rate (SR) and Success by weighted inverse Path Length (SPL) by a relative $\mathbf{3\%}$ and $\mathbf{1.3\%}$, respectively, compared to image-holistic baselines on the HM3D dataset validation split, while preserving interpretability and open-vocabulary flexibility. Code is available at: \href{https://uts-ri.github.io/RPV-SemNav/}{uts-ri.github.io/RPV-SemNav}.
\end{abstract}

\begin{comment}
\begin{IEEEkeywords}
Object-goal navigation, frontier exploration, open vocabulary, CLIP, semantic priors, co-occurrence, VLFM.
\end{IEEEkeywords}
\end{comment}

\section{Introduction}
The ability of an embodied agent to locate specific objects in a completely unknown environment, known as the Object-Goal Navigation (ObjectNav) problem~\cite{batra2020objectnav, ye2021auxiliary}, is foundational for general-purpose robotics. While geometric awareness provides the topological constraints necessary to navigate physical space, efficient task completion relies on semantic navigation, which infers spatial associations with the target from sparse visual cues. Thus, where geometry defines the navigable space, semantics prioritize the search space, transforming a random walk into an informed, goal-driven policy.

The current ObjectNav work can be divided into learning-based and zero-shot approaches. Learning-based methods learn semantic priors on a fixed list of object categories from large annotated indoor datasets~\cite{ramakrishnan2021hm3d, xia2018gibson, Matterport3D}, or apply reinforcement learning techniques for end-to-end navigation, while zero-shot approaches are open-vocabulary and generalizable, enabling navigation toward any target specified in natural language without task-specific fine-tuning. Recent advancements in these methods~\cite{yokoyama2023vlfm} leverage large-scale Vision-Language Models (VLMs)~\cite{radford2021clip, li2023blip2} for semantic inference, and combine this knowledge with explicit mapping and frontier selection~\cite{yamauchi1997frontier} for navigation. Many such approaches adopt a holistic image perspective, deriving a navigation signal by comparing a target string against the entire egocentric frame. We instead propose an object-centric zero-shot approach. By considering the environment to be a collection of semantically specific entities, an agent can derive search signals by treating detected objects as ``semantic anchors". While an object-centric approach offers more precise spatial grounding, a central issue remains that raw semantic similarity between object labels does not reliably encode spatial co-occurrence~\cite{chen2023semutil}. In the high-dimensional latent space of a VLM, linguistically and visually similar categories can exist close together, but be spatially unrelated.

\begin{figure}[t]
\centering
\includegraphics[width=\linewidth]{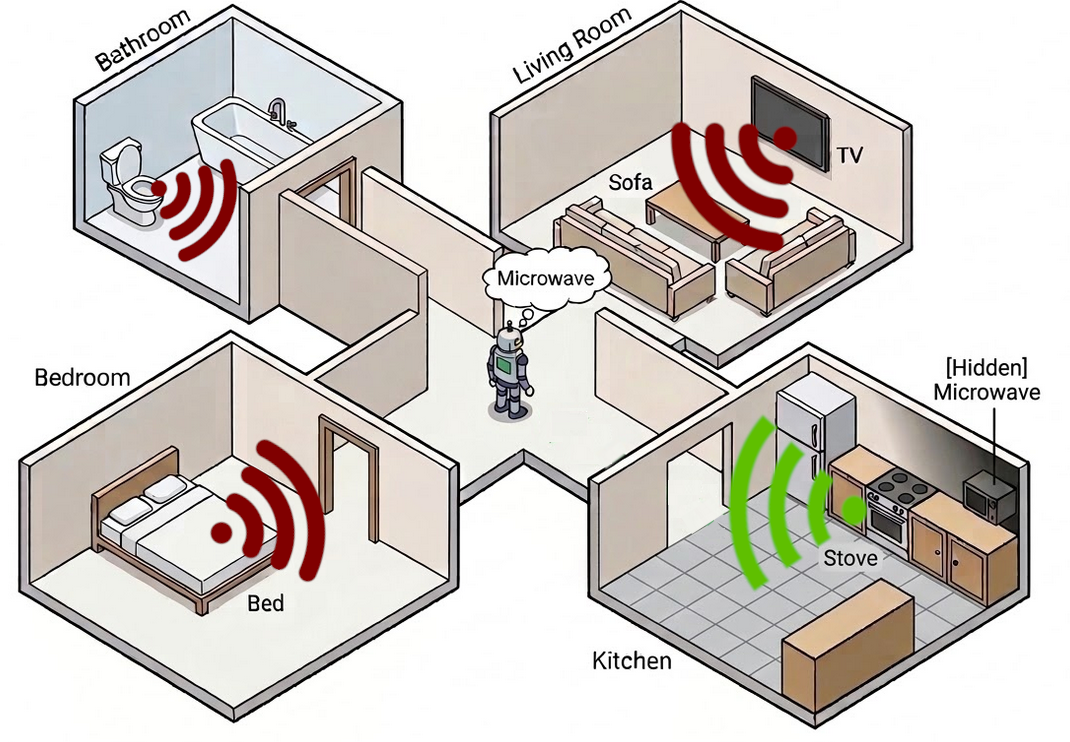}
\caption{Graphical overview of our proposed framework, example task of trying to find a microwave in a novel environment. Objects visible by the agent that have a high spatial relationship to the target emit stronger signals (indicated by a green signal symbol) than objects unlikely to share the same room type (indicated by a red signal symbol).}
\label{fig:overview}
\end{figure}

In this paper, we propose a novel analytical framework that bridges the gap between semantic similarity and spatial reality for the ObjectNav task. Rather than directly comparing object-to-object embeddings in the latent space, we develop a pipeline that mediates an object's semantic signal through a taxonomy of room types, thereby providing spatial context. Given a compact room lexicon, we map both the target and detected objects into a ``Room Probability Space", deriving a co-occurrence score reflecting the likelihood of any given object pair sharing the same location. To ensure these semantic priors respect the physical reality of the environment, we propagate object signals through free space using a geodesic flood-fill. This directly binds the semantic search space to the geometric constraints of the navigable space, scoring frontiers based on signal strength. The highest-scoring frontier is chosen as the next waypoint for navigation.

The main contributions of this work are as follows:
\begin{itemize}
  \item An analytical, room-mediated co-occurrence model that computes ``Room Probability Vectors" (RPVs) and estimates object-target spatial association from their probabilistic overlap.
  \item A geodesically constrained semantic value map formulation using flood-fill propagation with adaptive signal amplitude and spread.
  \item An integrated, object‑centric navigation pipeline supporting open‑vocabulary zero‑shot navigation with real‑time inference and no task‑specific fine‑tuning.
\end{itemize}

We demonstrate our pipeline on the HM3D dataset~\cite{ramakrishnan2021hm3d} within the Habitat simulator~\cite{savva2019habitat}, achieving competitive performance against prior state-of-the-art VLM zero-shot ObjectNav approaches, including relative improvements in the success rate ($+3\%$) and the success weighted by inverse path length ($+1.3\%$) metrics.

\section{Related Work}
\subsection{Learning-Based and End-to-End Navigation}
Learning-based approaches to the ObjectNav task have traditionally involved either learning semantic priors to guide exploration~\cite{chaplot2020semexp, ramakrishnan2022poni, luo2022stubborn} or using end-to-end techniques such as behavioral cloning~\cite{ramrakhya2023pirlnav} and reinforcement learning (RL)~\cite{ye2021auxiliary} for full autonomous navigation. Modular map-based pipelines such as SemExp~\cite{chaplot2020semexp} and PONI~\cite{ramakrishnan2022poni} isolate semantic mapping from the decision-making navigational process. These approaches train models on top-down maps to predict which unexplored regions are most likely to lead the agent toward the goal, effectively learning spatial layout priors and semantic co-occurrences from large-scale datasets. However, these methods are trained to only semantically understand a fixed set of predefined object categories. This restriction makes deployment into novel environments difficult, as the agent cannot generalize to open-vocabulary targets without in-domain fine-tuning or retraining. Alternatively, end-to-end frameworks such as PIRLNav~\cite{ramrakhya2023pirlnav} learn policies that implicitly convert observations into actions. Although effective, these methods incur high computational costs and often struggle with the sim-to-real domain gap when deployed in novel physical scenes.

\subsection{Image-Based Zero-Shot Navigation}
To overcome the limitations of closed-set vocabularies, recent works~\cite{yokoyama2023vlfm, bajpai2025uiapogn} exploit the visual-semantic representations inherent in foundational VLMs to enable open-vocabulary reasoning and zero-shot ObjectNav. CLIP on Wheels (CoW)~\cite{gadre2023cow} pioneered the use of CLIP~\cite{radford2021clip} text-image embeddings to relax constraints on target categories, allowing an agent to recognise and navigate towards arbitrary objects without task-specific training. VLFM~\cite{yokoyama2023vlfm} computes language-grounded value maps to guide frontier selection~\cite{yamauchi1997frontier} by calculating cosine similarity between BLIP-2~\cite{li2023blip2} image embeddings and handcrafted text prompts. UIAP-OGN~\cite{bajpai2025uiapogn} addresses prompt sensitivity by ensembling multiple prompts to model linguistic uncertainty, but this increases inference cost.

A common limitation of \emph{VLFM-style image-holistic methods} is the use of visibility-cone projection for semantic mapping. In these pipelines, semantic evidence is primarily updated from the agent’s current line of sight, so value maps can remain sparse and concentrated in the observed regions. As a result, frontiers outside the current field of view may receive delayed semantic updates, even when they are geometrically reachable.

\subsection{Object-Centric and Language-Guided Navigation}
Several \emph{zero-shot} object-goal navigation methods move beyond image‑holistic scoring by leveraging object-level cues. L3MVN~\cite{yu2023l3mvn} and ESC~\cite{zhou2023esc} construct ``search windows" around frontiers, extract visible objects, and use Large Language Model (LLM) reasoning to rank candidate frontiers. LOAT~\cite{lin2025loat} reduces runtime LLM queries by \emph{a priori} extracting commonsense object relations, but still depends on a learned experience module, thus retaining dependence on task-specific training signals.

Other approaches embed object labels to guide frontier selection. SemUtil~\cite{chen2023semutil} builds spatial scene graphs by embedding object labels near each frontier with BERT~\cite{devlin2019bert}, while SEEK~\cite{ginting2024seek} extends this to room-level reasoning using trained relational semantic networks. However, direct language-space similarity can be an unreliable proxy for physical co-occurrence, and learned relational modules introduce additional data and training assumptions. This highlights the need for a zero-shot, training-free framework that grounds object semantics into room-structured spatial priors for unmapped environments.

\section{Method}
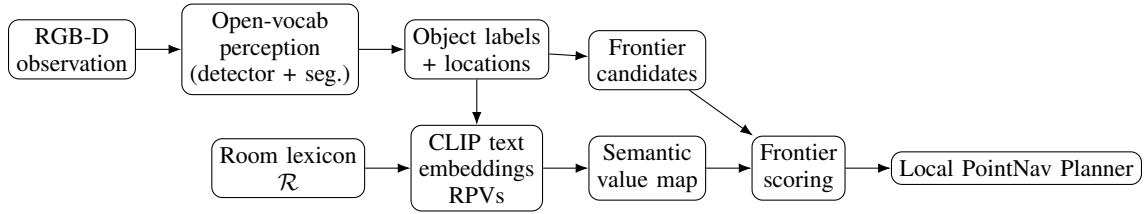
\begin{figure*}[t]
\centering
\begin{tikzpicture}[node distance=6mm, font=\small,
    box/.style={draw, rounded corners, align=center, inner sep=3pt}]
\node[box] (rgbd) {RGB-D\\observation};
\node[box, right=of rgbd] (percep) {Open-vocab\\perception\\(detector + seg.)};
\node[box, right=of percep] (objects) {Object labels\\+ locations};
\node[box, below=of objects] (rpv) {CLIP text\\embeddings\\RPVs};
\node[box, left=of rpv] (rooms) {Room lexicon\\$\mathcal{R}$};
\node[box, right=of rpv] (value) {Semantic\\value map};
\node[box, above=of value] (frontier) {Frontier\\candidates};
\node[box, right=of value] (score) {Frontier\\scoring};
\node[box, right=of score] (plan) {Local PointNav Planner};
\draw[-{Latex}] (rgbd) -- (percep);
\draw[-{Latex}] (percep) -- (objects);
\draw[-{Latex}] (objects) -- (frontier);
\draw[-{Latex}] (objects) -- (rpv);
\draw[-{Latex}] (rooms) -- (rpv);
\draw[-{Latex}] (rpv) -- (value);
\draw[-{Latex}] (frontier) -- (score);
\draw[-{Latex}] (value) -- (score);
\draw[-{Latex}] (score) -- (plan);
\end{tikzpicture}
\caption{Overview of the proposed system: main novelty is the deterministic room-mediated co-occurrence scoring (RPVs) and its use in frontier scoring.}
\label{fig:pipeline}
\end{figure*}

A high-level overview of the system is shown in Fig.~\ref{fig:pipeline}. The novelty is in computing room-mediated co-occurrence priors and injecting them into frontier scoring.

\begin{figure*}[t]
\centering
\setlength{\tabcolsep}{2pt}
\begin{tabular}{ccc}
\includegraphics[width=0.32\textwidth]{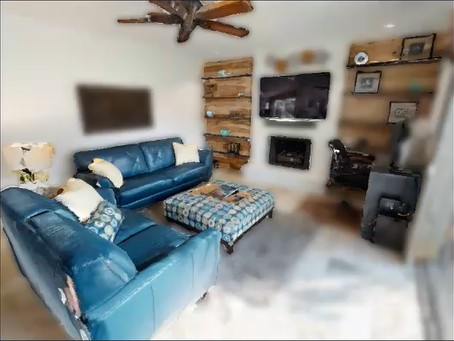} &
\includegraphics[width=0.32\textwidth]{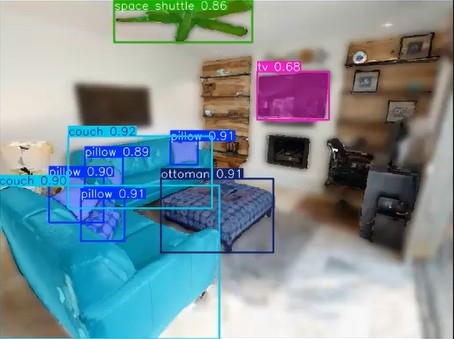} &
\includegraphics[width=0.32\textwidth]{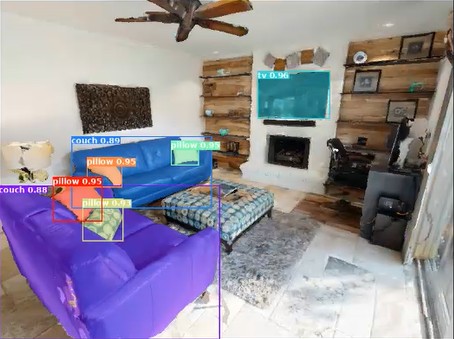} \\
\small \textbf{(a)} Mask and blur irrelevant regions &
\small \textbf{(b)} YOLO detections (boxes, labels) &
\small \textbf{(c)} SAM masks (pixel-accurate segments)
\end{tabular}
\caption{Perception pipeline used to build an object-centric map while suppressing distractions. (a) Regions and classes not used for navigation are masked and blurred to reduce spurious detections. (b) A detector proposes object candidates with bounding boxes and confidence scores. (c) The labels serve as prompts for a segmentation model to produce refined instance masks, which are subsequently used for semantic scoring.}
\label{fig:perception_pipeline}
\end{figure*}

\subsection{Task Definition}
The ObjectNav~\cite{batra2020objectnav} task requires a robot agent to explore an unknown environment, given an arbitrary starting pose, to locate a specified target object. The robot operates in three modes: \{\text{initialize}, \text{explore}, \text{navigate}\}. In ``initialize", the robot performs 12 left turns to survey an entire $360^\circ$ view of its starting location. During ``explore", it captures an egocentric RGB-D observation at each timestep. It executes an action $a_t \in \mathcal{A}$ from the discrete set: 
{
\[
\mathcal{A}=\{\text{forward}(0.25\text{m}),\,\text{left}(30^\circ),\,\text{right}(30^\circ),\,\text{stop}\}.
\]
}

Robot odometry is provided by the Habitat simulator~\cite{savva2019habitat}. When the target object is detected, the robot goes into ``navigate" and solely focuses on reaching the goal. An episode succeeds if the robot navigates to within $1$m of a target instance in less than $T=500$ steps. Unlike methods trained on fixed category sets, our open-vocabulary approach handles arbitrary object categories expressible in natural language.

\subsection{Perception Pipeline}
The perception pipeline extracts object instances for mapping and computing semantic relevance scores, as shown in Fig.~\ref{fig:perception_pipeline}.

\textbf{Background Masking.}
We first mask image regions corresponding to categories that do not contribute to object-goal navigation, such as structural surfaces. Semantic segmentation is performed using Mask2Former~\cite{cheng2022mask2former} and a model trained on the ADE20K~\cite{zhou2017ade20k} dataset with a ResNet-50~\cite{he2016resnet} backbone. The ADE20K class list defines each category as either ``thing" or ``stuff", where ``stuff" refers to architectural instances such as ``floor", ``wall", and ``ceiling". These segments are blurred to prevent them from generating detections downstream. Although ADE20K does not classify ``door", ``window", or ``curtain" as ``stuff", we blur these categories as well, as they contribute little semantic discrimination for navigation and often introduce noise.

\textbf{Open-Vocabulary Detection + Prompt Proposal.}
Next, an open-vocabulary YOLOE~\cite{wang2025yoloe} detector is applied on the masked image with an unfiltered list of 1203 categories defined in the LVIS dataset~\cite{gupta2019lvis}. YOLOE is chosen for its open-vocabulary nature and computational efficiency. All detections with confidence above 0.5 are retained as proposed prompts for a subsequent high-accuracy segmentation stage. The broad, unrestricted LVIS category set makes this step sensitive to false positives, underscoring the importance of the initial masking stage.

\textbf{High-Accuracy Segmentation.}
We cross-verify the list of detections by feeding the labels as prompts to SAM 3~\cite{carion2025sam3}, a high-accuracy segmentation model. This step filters out false positives, preventing undesirable signals from permeating through the map. SAM 3 returns pixel-accurate instance masks, from which object centroids are extracted and projected into world coordinates via depth information. The target category is always included in the prompt list to enable immediate goal recognition.

\subsection{Room-Mediated Co-occurrence Model}
\subsubsection{Room Lexicon}
Object co-occurrence in indoor environments is closely tied to functional context. Objects appear near one another when serving related or synergistic purposes. Room types, therefore, provide a natural abstraction for semantically grouping objects and imposing the spatial context necessary for reasoning about co-occurrence. We define a compact room lexicon $\mathcal{R}$ that serves as an intermediary semantic space for estimating spatial association between objects. For indoor spaces, focusing primarily on household environments, we use a variation on the list of room types in the HM3D dataset~\cite{ramakrishnan2021hm3d}: \{\emph{bathroom}, \emph{bedroom}, \emph{dining room}, \emph{garage}, \emph{kitchen}, \emph{hallway}, \emph{laundry}, \emph{living room}, \emph{home office}\}. Importantly, $\mathcal{R}$ is not fixed; users may substitute domain-specific or more abstract categories (e.g. ``sleeping area", ``cooking area", etc.) without requiring retraining, enabling zero-shot spatial reasoning.

\subsubsection{Room Probability Vectors (RPVs)}
Let the room lexicon $\mathcal{R}=\{r_1,\dots,r_K\}$ and denote $e(\cdot)$ as the CLIP text embedding of a label. To avoid sensitivities inherent to prompt engineering, we use raw class names without template prompts. For an object $o$, we compute its similarity to each room label via cosine similarity:
\begin{equation}
s_o(r_i) = \cos\left(\text{e}(o), \text{e}(r_i)\right) 
         = \frac{\text{e}(o) \cdot \text{e}(r_i)}{\|\text{e}(o)\| \, \|\text{e}(r_i)\|}
\label{eq:clip_similarity}
\end{equation}

A room-probability distribution is then obtained by applying a temperature-scaled softmax function:

\begin{equation}
p_o(r_i) = \frac{\exp(s_o(r_i)/\tau)}{\sum_{j=1}^K \exp(s_o(r_j)/\tau)}
\label{eq:softmax}
\end{equation}
where $\tau$ controls the sharpness of the room-probability distribution. We set $\tau=0.03$ to balance discriminative power and robustness to noise in CLIP's embeddings. Lower temperatures lead to over-confident distributions, whereas higher values yield under-informative, near-uniform distributions. The \emph{Room Probability Vector} (RPV) $\mathbf{p}_o = [p_o(r_1), \ldots, p_o(r_K)]^\top$ represents the estimated likelihood of an object $o$ being associated with each room type.

\subsubsection{Room-Mediated Co-occurrence Scoring}
We treat the RPV $\mathbf{p}_o$ as the parameters of a categorical distribution for a random variable ${X}_o$, such that $P({X}_{o}={r}_{i})={p}_{o}({r}_{i})$. To estimate the co-occurrence between a detected object $o_d$ and the target object $o_t$, we compute the dot product of their respective RPVs:
\vspace{-2pt}
\begin{equation}
{A}_{o} = \text{cooc}(o_d, o_t) = \mathbf{p}_{o_d} \cdot \mathbf{p}_{o_t} 
                       = \sum_{i=1}^K p_{o_d}(r_i) \cdot p_{o_t}(r_i)
\label{eq:cooccurrence}
\end{equation}

This operation yields the \emph{Joint Probability of Co-occurrence}, $P({X}_{o_d}={X}_{o_t})$, representing the likelihood of objects being associated with the same room categories, thus sharing spatial context. By marginalizing over the latent room basis $\mathcal{R}$, this formulation provides a probabilistic grounding rather than relying on raw embedding similarity. Consequently, object pairs with significant distributional overlap receive high scores, while functionally disjoint pairs are naturally suppressed.

\subsection{Signal Propagation}
\subsubsection{Geometrically-Aware Value Mapping}
Each detected object will emit a spatially decaying signal into a value map to guide the agent through navigation. We perform a flood-fill using the Fast Marching Method (FMM)~\cite{sethian1999fmm} in combination with a continuously evolving occupancy grid to ensure that signals respect the topology of the navigable space. Let $\mathcal{O}_{\mathbf{x}}$ denote the set of detected objects whose signals reach cell location $\mathbf{x}$. The value at $\mathbf{x}$ aggregates these contributions using normalized summation to prevent object-density bias:

\begin{equation}
V(\mathbf{x}) = \frac{1}{|\mathcal{O}_{\mathbf{x}}|}\sum_{o\in\mathcal{O}_{\mathbf{x}}} {A}_{o} \cdot \exp \left(-\frac{{d}_{geo}(\mathbf{x}, \mathbf{x}_o)^2}{2\sigma_o^2}\right)
\label{eq:fmm}
\end{equation}
where $\mathcal{O}_{\mathbf{x}} \subseteq \mathcal{O}$, ${d}_{geo}$ is the geodesic distance through free space from object centroid position $\mathbf{x}_o$ to cell position $\mathbf{x}$, and ${A}_{o}$, $\sigma_o$ are signal amplitude and spatial spread parameters (\S\ref{sec:adaptive_params}).
Using FMM for flood-fill guarantees that signals do not penetrate through walls, thereby preventing any influence on frontiers in adjacent, unrelated rooms, a failure mode of Euclidean distance-based approaches. We validate this design choice against isotropic Gaussian propagation in Section~\ref{sec:signal_ablation}. Normalization ensures cells affected by multiple relevant objects receive compound evidence while maintaining the $[0,1]$ range, preventing regions with many weak detections from dominating regions with few strong detections.

\subsubsection{Adaptive Sigma} \label{sec:adaptive_params}
We modulate the signal decay rate of the detected objects based on the calculated co-occurrence score relative to the target object. Objects of high relevance should guide exploration across a broader area, whereas weak associations should exert only local influence. 
We apply a square root transformation to the signal amplitude such that:

\begin{equation}
{w}=\sqrt{A_o}
\label{eq:amplitude}
\end{equation}

The resulting value ${w}$ is used to determine the rate of decay. Observing that a large distribution of scores clusters within the $[0.1, 0.3]$ range, as displayed in Fig.~\ref{fig:rpv_heatmap}, we apply this correction to improve the discriminatory power of objects whose co-occurrence score falls in the upper echelon of this band. This comes at the cost of compressing the overall dynamic range. However, instances of extremely relevant objects are still scored highly, and the extent of their relevance is further boosted, making the trade-off worthwhile. 

The spatial spread of the signal scales proportionally with the weighting factor, so that high-relevance objects exert a wider influence and low-relevance objects have a more local effect. We define $\sigma_o$ as a linear interpolation between minimum and maximum values:
\begin{align}
\sigma_o &= \sigma_{\min} + (\sigma_{\max} - \sigma_{\min}) \cdot {w}\label{eq:adaptive_sigma}
\end{align}
where $\sigma_{\min} = 1.0\,$m and $\sigma_{\max} = 5.0\,$m. This ensures high-relevance objects guide navigation over room-scale distances, while low-relevance objects have only localized influence.

\subsection{Frontier Selection}
At each timestep, we extract frontiers, defined as boundaries between free and unexplored space, and cluster neighboring free space cells within a radius $r = 0.5\,$m into candidate regions $\mathcal{F} = \{F_1, F_2, \ldots, F_N\}$, where $\mathcal{C}_j$ denotes the set of free-space cells comprising region $F_j$. Similar to VLFM~\cite{yokoyama2023vlfm}, for each region $F_j \in \mathcal{F}$, the median value is calculated and used as the frontier score:

\begin{equation}
V_j = \underset{c \in \mathcal{C}_j}{\text{median}}(V(c))
\label{eq:frontier_score}
\end{equation}
where $V(c)$ is given by Eq.~\eqref{eq:fmm}. This captures the representative signal strength in the vicinity of the frontier. The highest scoring frontier is selected for navigation: 
\begin{equation}
F^* = \operatorname*{arg\,max}_{F_j \in \mathcal{F}} V_j
\label{eq:frontier_selection}
\end{equation}

Given a selected goal or frontier waypoint, the robot is driven by a fixed, pretrained PointNav controller shared across methods (following the VLFM-based implementation), so differences arise from frontier scoring rather than low-level control.

\begin{figure*}[t]
\centering
\setlength{\tabcolsep}{2pt}
\begin{tabular}{ccc}
\includegraphics[width=0.32\textwidth]{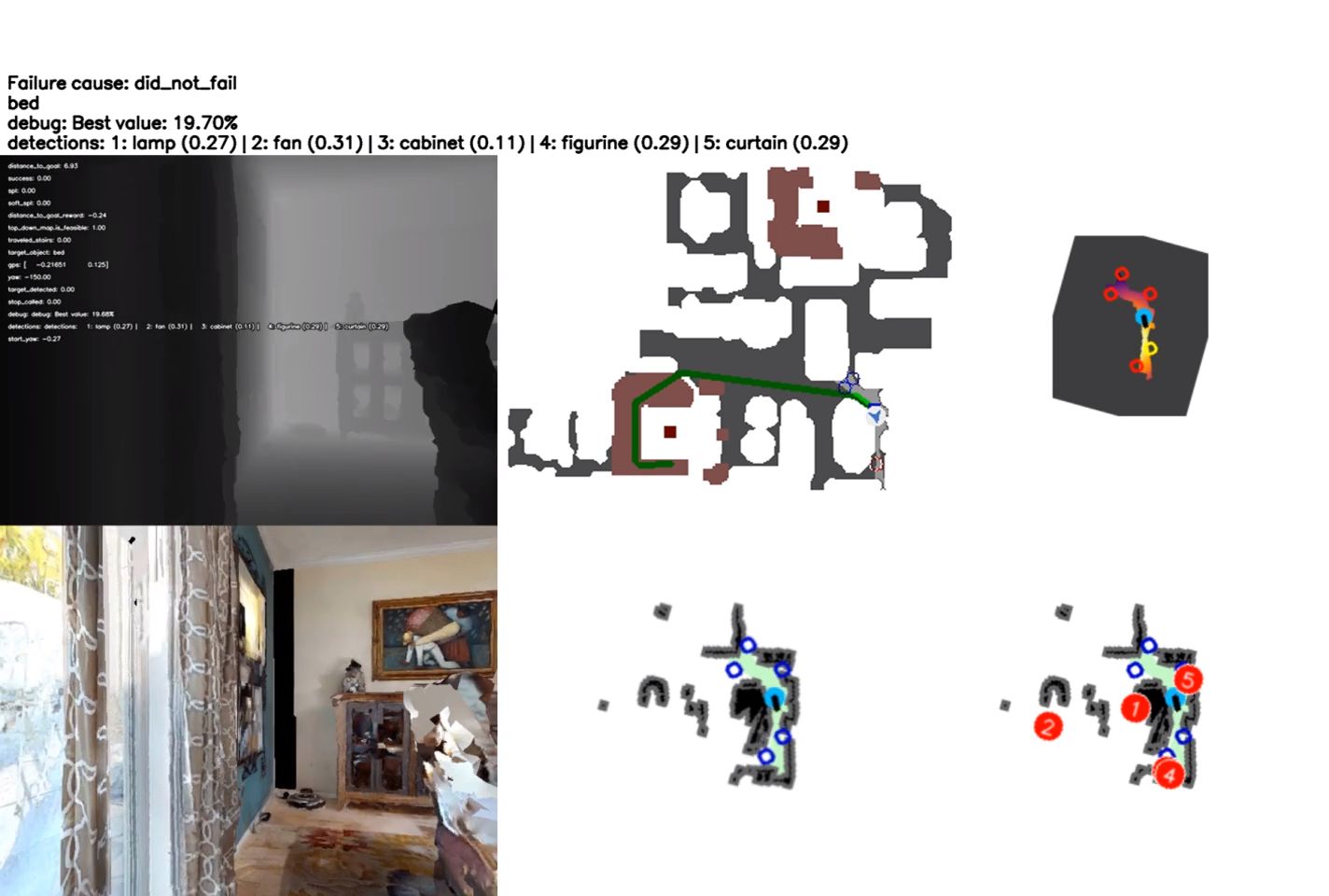} &
\includegraphics[width=0.32\textwidth]{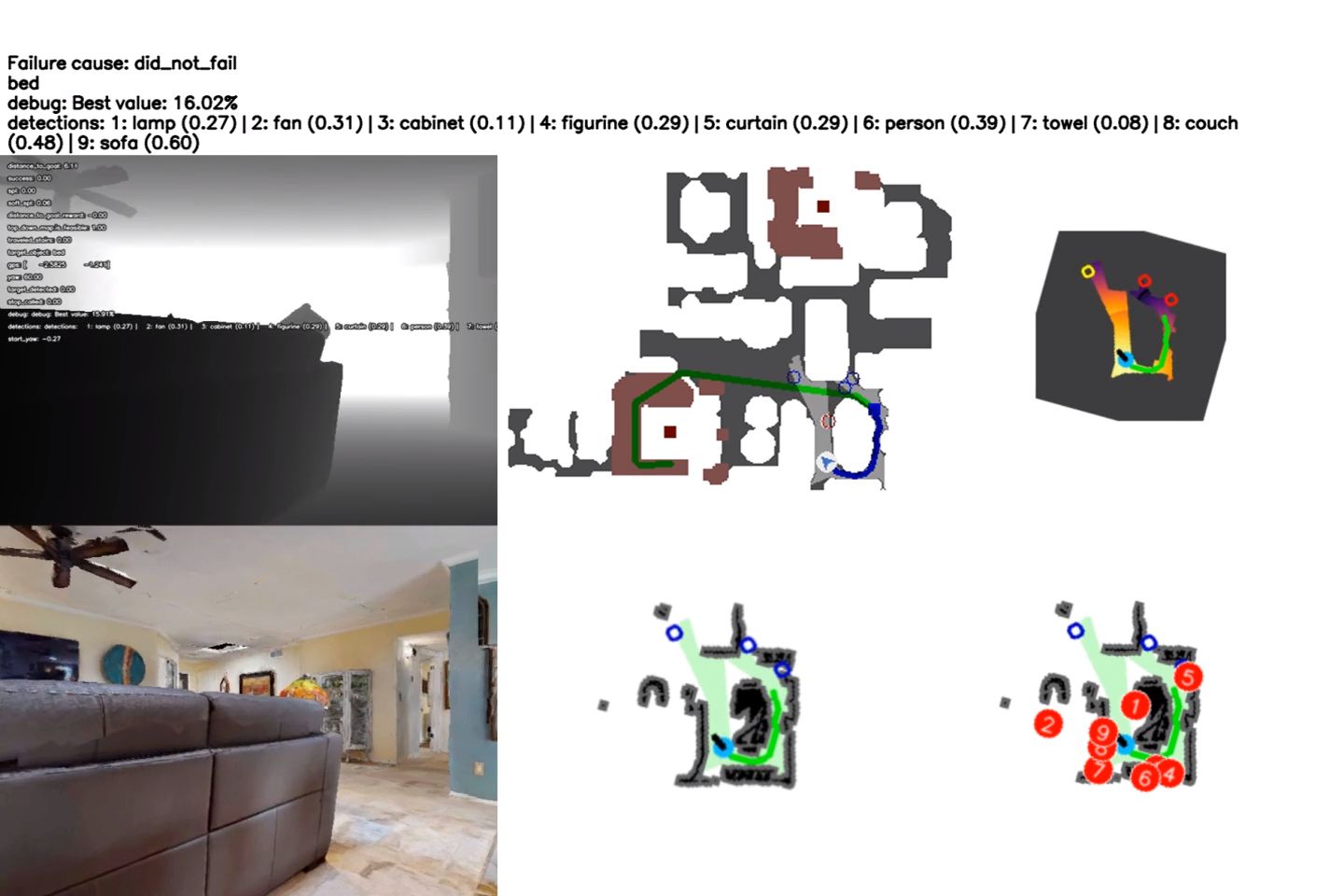} &
\includegraphics[width=0.32\textwidth]{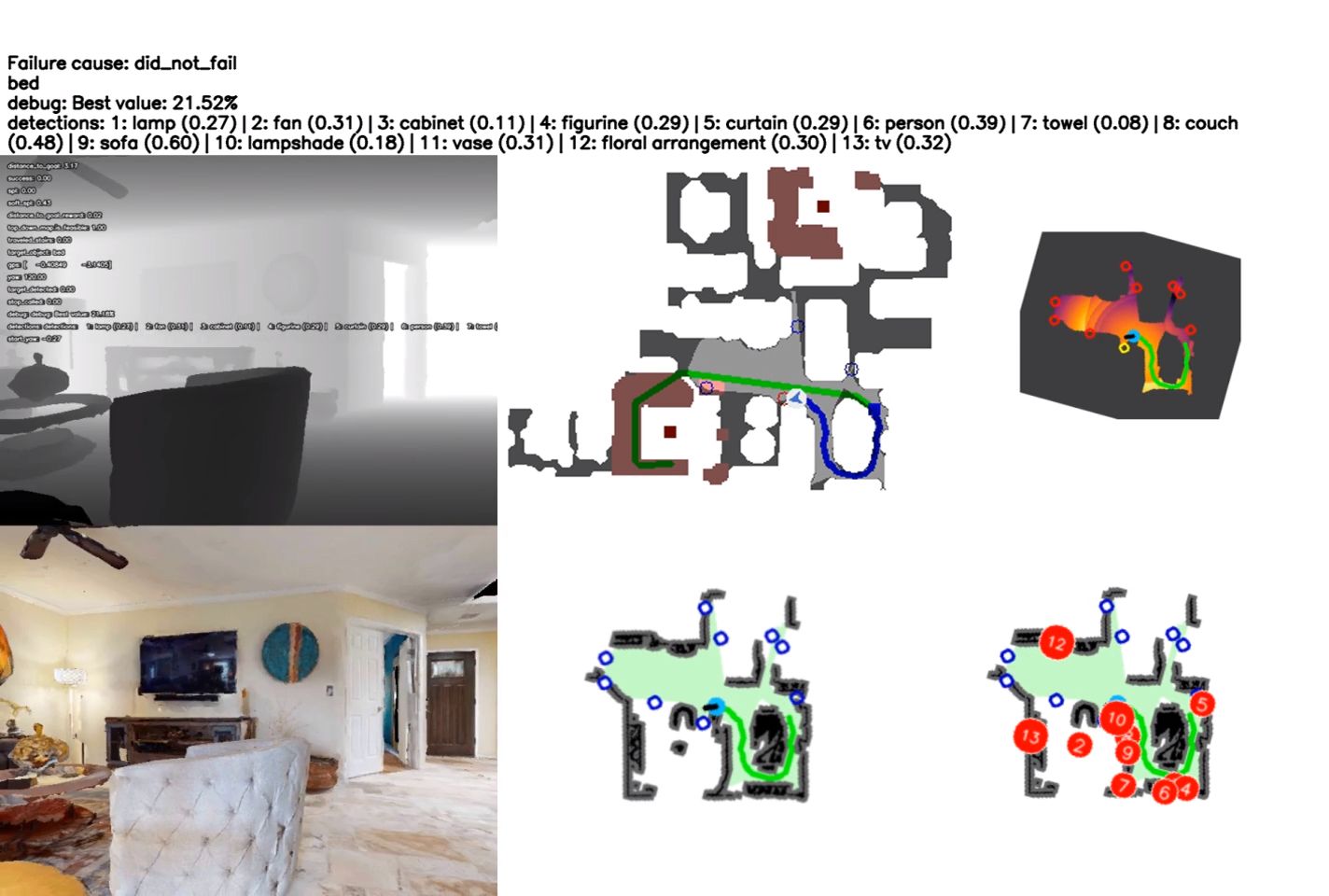} \\
\small \textbf{(a)} Early exploration &
\small \textbf{(b)} Expanding mapped area &
\small \textbf{(c)} Semantic focus increases \\
\includegraphics[width=0.32\textwidth]{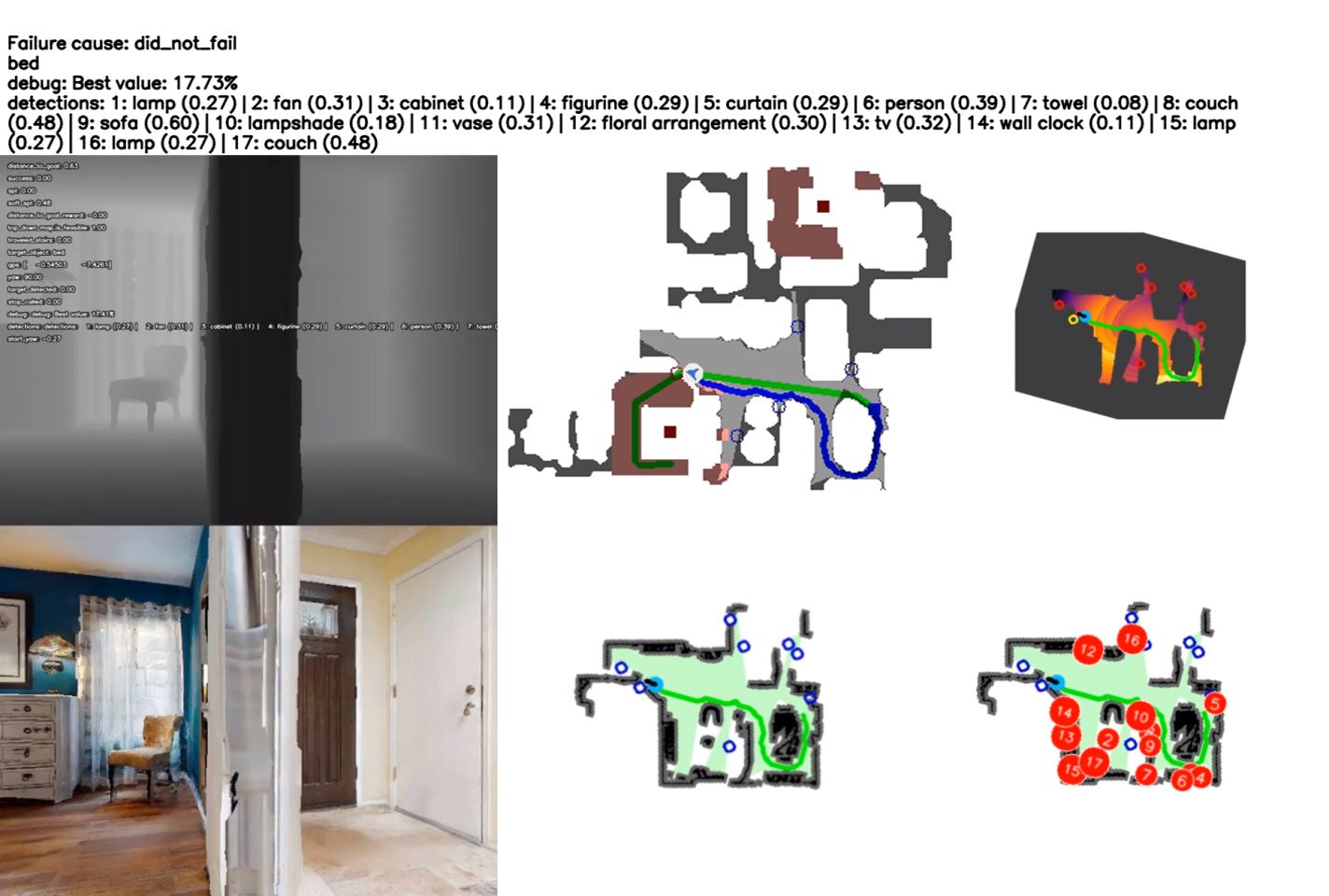} &
\includegraphics[width=0.32\textwidth]{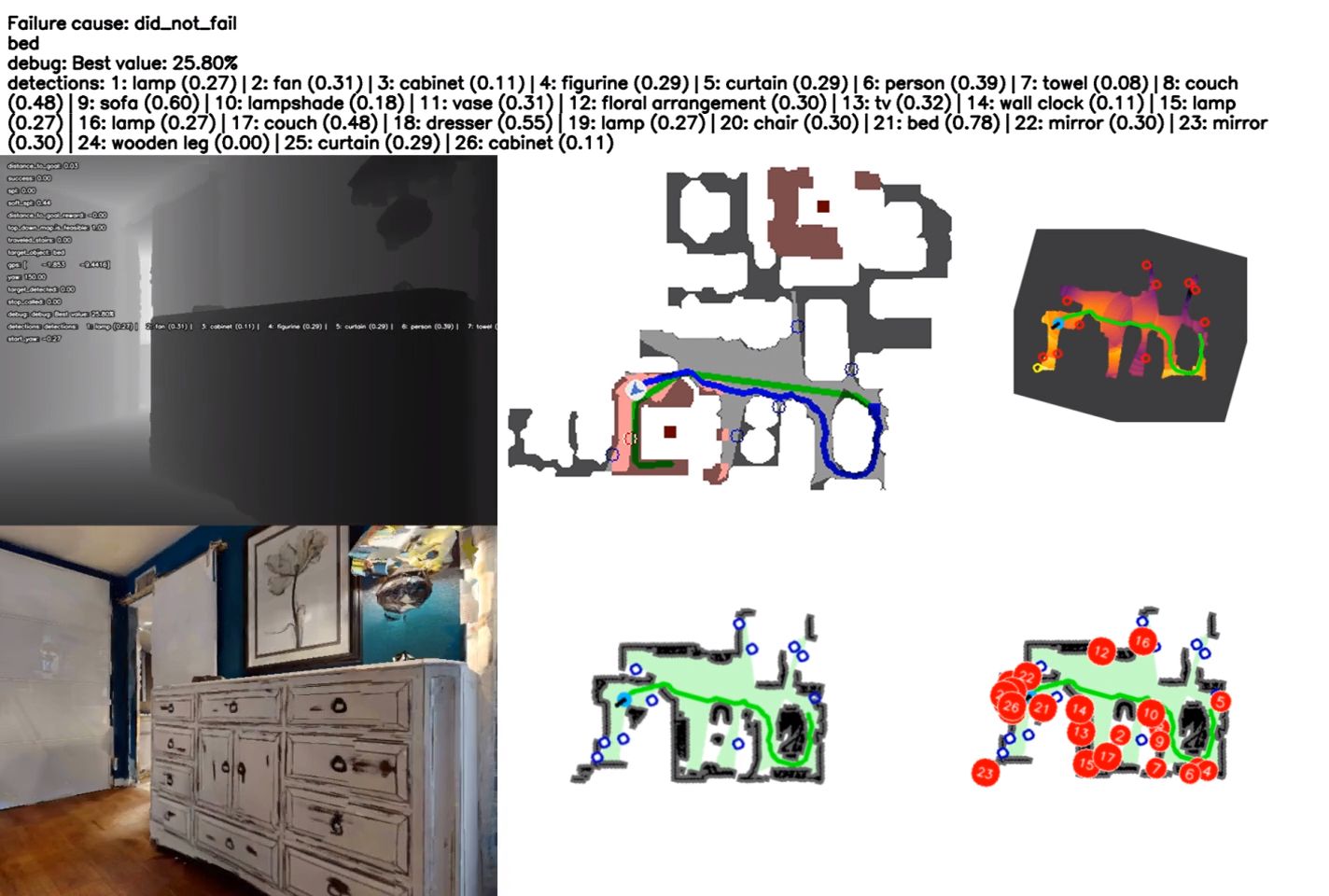} &
\includegraphics[width=0.32\textwidth]{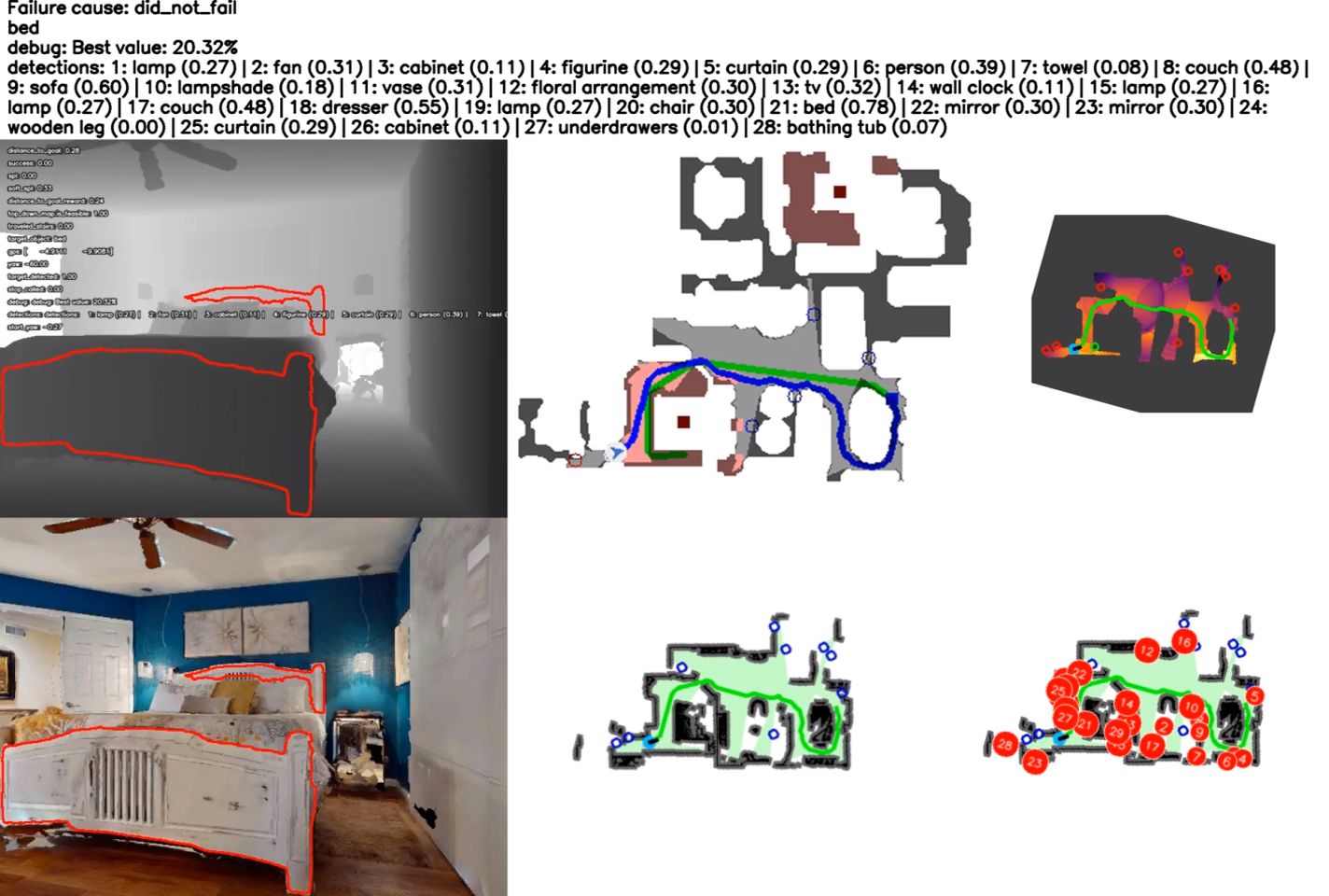} \\
\small \textbf{(d)} Frontier refinement &
\small \textbf{(e)} Room transition &
\small \textbf{(f)} Bed region reached
\end{tabular}
\caption{Qualitative progression of one ObjectNav episode with target class \emph{bed}. Across the six snapshots, the agent incrementally expands free-space coverage (map panel), updates semantic value around plausible goal regions (heatmap panel), and refines frontier selection as new observations are integrated. The final frame shows successful visual grounding of the bed region while the explored trajectory and accumulated frontier evidence converge near the goal room.}
\label{fig:qualitative}
\end{figure*}

\section{Experimental Setup}
\subsection{Simulation, Datasets and Hardware}
\label{sec:setup}
We evaluate in the Habitat~\cite{savva2019habitat} simulator on the validation split of the HM3D semantics dataset~\cite{yadav2022hm3dsemantic}. This consists of 2000 unique episodes split over 20 indoor scenes and six defined target objects $\mathcal{T}=\{\emph{chair}, \emph{bed}, \emph{potted plant}, \emph{toilet}, \emph{tv}, \emph{couch}\}$. All evaluations were run on a PC with a 12th Gen Intel Core i9-12900K CPU and an NVIDIA Quadro RTX 8000 GPU with 48GB of VRAM.

\subsection{Metrics and Baselines}
\label{sec:metrics_and_baselines}
We evaluate two versions of our pipeline, \textbf{RPV-V1} and \textbf{RPV-V2}, to isolate the impact of our signal-propagation method. In \textbf{RPV-V1}, we model signals as isotropic Gaussian distributions stamped onto the value map, which strictly decay with Euclidean distance and neglect environmental topology. In \textbf{RPV-V2}, we use FMM flood-fill to ensure that signals propagate geodesically and do not permeate physical walls.

We report Success Rate (SR), Success weighted by inverse Path Length (SPL), and Distance to Goal (DTG) in meters~\cite{anderson2018agents} for navigation. 
To contextualize our performance, we compare against prominent zero-shot and learning-based baseline approaches. For zero-shot, training-free methods, we evaluate against VLFM~\cite{yokoyama2023vlfm}, the state-of-the-art, and UIAP-OGN~\cite{bajpai2025uiapogn}, which extends VLFM by ensembling results across multiple input prompts, incorporating linguistic uncertainty into value mapping, and reducing reliance on prompt engineering. As an LLM-incorporated method, we also compare against the zero-shot variant of L3MVN~\cite{yu2023l3mvn}. To establish the performance ceiling of methods requiring in-domain training, we also compare against a trained end-to-end policy in PIRLNav~\cite{ramrakhya2023pirlnav} and the feed-forward variant of L3MVN~\cite{yu2023l3mvn}.

\subsection{Perception Throttling}
In our pipeline, we run the full perception stack once every three steps while the robot is in ``explore" mode. Once the target object is found, we run only SAM 3 segmentation once every three steps in ``navigate" mode (masking and detection are unnecessary). Given discrete robot actions and typical indoor scene layouts, consecutive frames rarely introduce novel objects, so per-step masking and detection are redundant and increase computation.

\subsection{Ground-Truth Conditional Probability}
To evaluate the fidelity of our Room Probability Vectors independent of navigation performance, we establish a ground-truth co-occurrence benchmark derived from annotations in the HM3D semantic dataset~\cite{yadav2022hm3dsemantic}. We compute a matrix, $P(t\mid o)$ representing the conditional probability of encountering target object $t$ in the same room as detected object $o$:
\begin{equation}
P(t\mid o)=\frac{|\mathcal{R}_t \cap \mathcal{R}_o|}{|\mathcal{R}_o|}
\label{eq:cond_prob}
\end{equation}
where $\mathcal{R}_t$ is the set of rooms containing target object $t$ and $\mathcal{R}_o$ is the set of rooms containing detected object $o$.

When computing RPVs and conditional probabilities, we use the Matterport semantic mapping table defined in Habitat~\cite{savva2019habitat} to map raw, noisy object labels to 40 standardized categories. Following VLFM~\cite{yokoyama2023vlfm} and PONI~\cite{ramakrishnan2022poni}, we exclude non-discriminative architectural categories such as \emph{floor}, \emph{wall}, and \emph{ceiling}, which appear abundantly across all room types, providing negligible information for spatial reasoning. This filtering yields a set of 21 object classes, including the subset of target objects $\mathcal{T}$ defined in Section~\ref{sec:setup}. For each target category, we compute Spearman's rank correlation ($\rho$) between RPV comparison measures and ground-truth co-occurrence, evaluated both across the full candidate set and the ground truth Top-5 anchors, alongside Recall@5 and Normalized Discounted Cumulative Gain (NDCG@5).

\section{Results}

\subsection{Co-occurrence Measure Validation}
\begin{figure}[t]
\centering
\includegraphics[width=\linewidth]{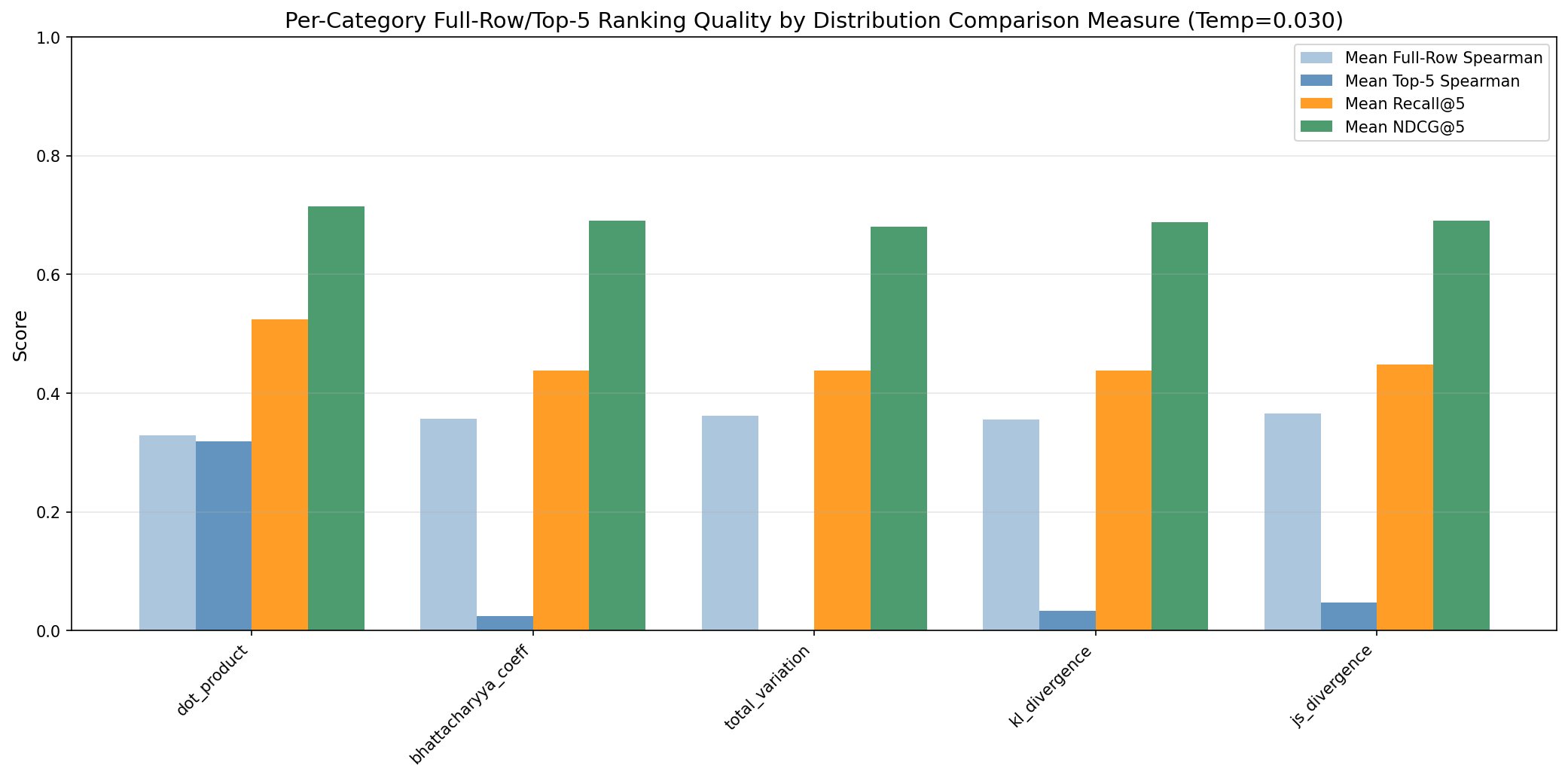}
\caption{Per-category rank agreement between distribution comparison measures and HM3D MP21 categories ground-truth co-occurrence, reported as full-row Spearman, Top-5 Spearman, Recall@5 and NDCG@5 at $K=5$.}
\label{fig:metric_correlation}
\end{figure}

Fig.~\ref{fig:metric_correlation} shows the mean full-row and Top-5 per-category Spearman rank correlation ($\rho$) between different Room Probability Distribution comparison measures and the conditional probability ground truth, at softmax temperature $\tau=0.03$. Mean Recall@5 and NDCG@5 were also computed for each measure. The RPV dot product exhibited the highest Top-5 Spearman rank correlation ($\rho=0.319$), Recall@5 ($0.524$) and NDCG@5 ($0.715$), making it the strongest proxy for spatial association despite a slightly lower full-row Spearman correlation relative to the other four measures ($\rho=0.329$ compared to $\rho=0.355-0.365$).

For the ObjectNav task, the relative ranking of object associations in the ``Room Probability Space" is more critical than their absolute alignment. From an object-centric perspective, given a target object $t$ and a set of detected objects $\mathcal{O}$, the agent must prioritize navigation toward anchor objects $o$ in an order that reflects the likelihood of physical co-occurrence. Correlation is computed per target category rather than globally, reflecting how the navigation signal during inference is taken as the comparison of a fixed target RPV against those of detected objects. Ranking quality is evaluated at $K=5$ as the ground truth co-occurrence distribution is concentrated among a small number of strong anchors out of the 21 candidate categories per target. Correctly prioritizing this subset is what determines navigation efficiency, as the relative order of weakly-associated candidates carries comparatively little operational consequence. Although the Spearman correlation is modest, the Recall@5 and NDCG@5 scores, respectively, indicate that high co-occurrence anchors are reliably surfaced by the RPV dot product, and of those anchors, the strongest associated objects are commonly assigned the greatest weight. This is notable given the method's zero-shot nature and the domain gap between VLM text embeddings and physical 3D environmental layouts.

\begin{figure}[t]
\centering
\includegraphics[width=\linewidth]{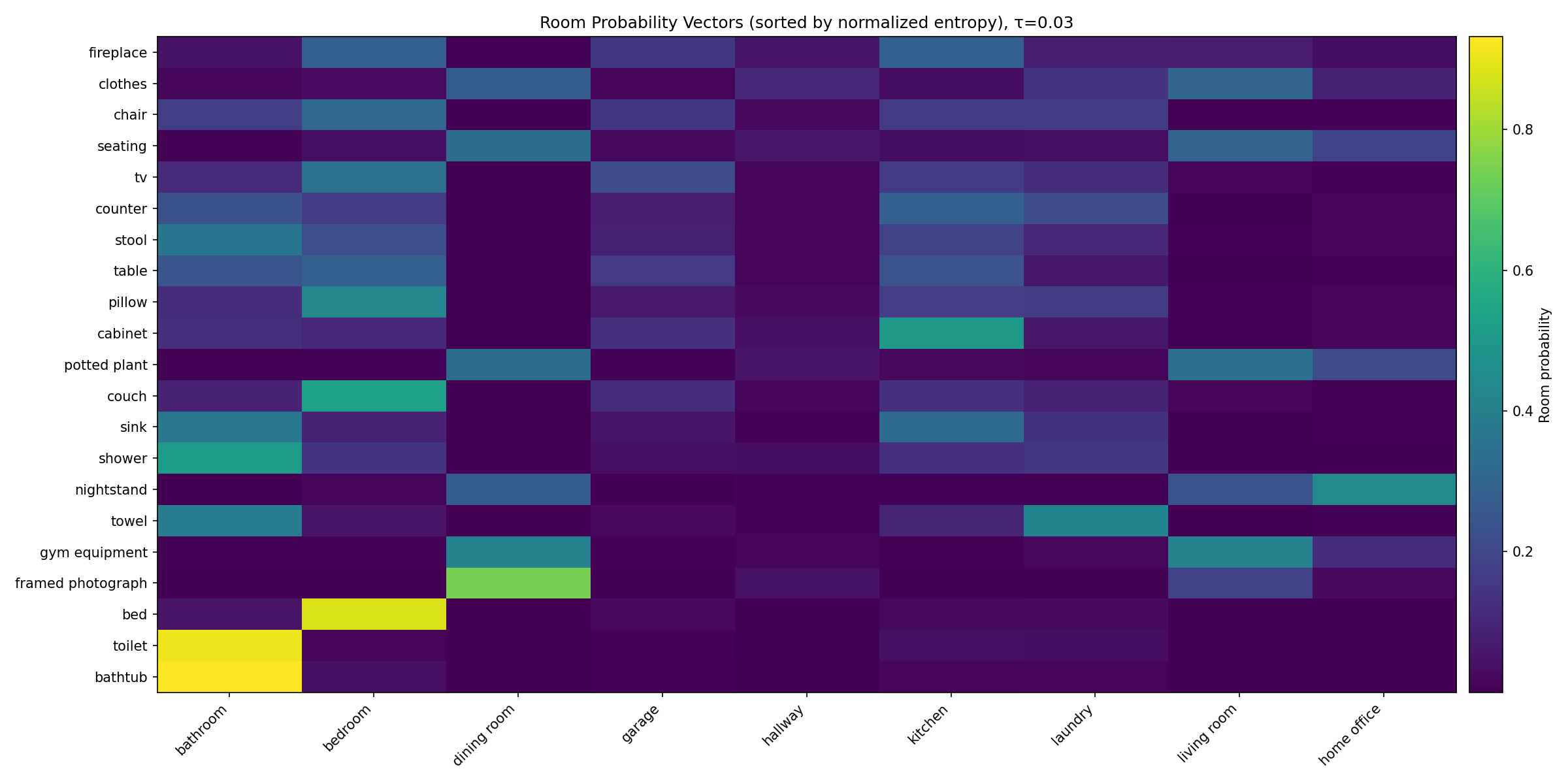}
\caption{Room Probability Vectors (RPVs) sorted by normalized entropy (top: most uniform, bottom: most peaked). Low-entropy objects (e.g., bed, bathtub, toilet) are strongly concentrated in a single room type, whereas high-entropy objects (e.g., fireplace, clothes, chair) are spread across many rooms, indicating weak spatial specificity.}
\label{fig:rpv_heatmap}
\end{figure}

Although the dot product is a symmetric measure, the dataset's conditional distribution $P(t\mid o)$ is inherently asymmetric. Notably, it nonetheless outperforms KL divergence, the only directed measure evaluated. More broadly, we hypothesize the remaining measures are more sensitive to the high normalized entropy of ``generic" objects (e.g. \emph{chair}: $\hat{H}=0.773$), whose probability mass is distributed across many room types, visualized in Fig.~\ref{fig:rpv_heatmap}, as none share the multiplicative suppression trait of the dot product. Conversely, low-entropy structural fixtures tightly concentrate their mass, serving as reliable anchors. The dot product, while influenced by entropy, mitigates it by measuring mutual overlap rather than directional divergence.

However, this suppression of low-magnitude mass, while advantageous for isolating peak alignment, explains the convergence in full-row Spearman correlation (Fig.~\ref{fig:metric_correlation}). As the candidate set expands beyond the strongest objects, the distribution of high-entropy RPVs (Fig.~\ref{fig:rpv_heatmap}) causes dot product scores to collapse into a narrow, low‑magnitude band, making genuine associations difficult to distinguish from noise. This highlights the necessity of the non-linear square root transformation in Eq.~\ref{eq:amplitude}. By expanding the relative differences between low-scoring pairs, the transformation ensures semantically meaningful but weaker co‑occurrence relationships remain detectable during navigation.

\begin{table}[t]
\centering
\caption{HM3D ObjectNav Performance. $\dagger$ denotes baselines re-evaluated locally. \textbf{Bold} denotes best overall metric. \underline{Underscore} denotes best from non-trained approaches.}
\label{tab:objectnav_results}
\begin{tabular}{lcccc}
\toprule
Method & Trained? & SR $\uparrow$ & SPL $\uparrow$ & DTG $\downarrow$ \\
\midrule
L3MVN (feed-forward)~\cite{yu2023l3mvn} & Yes & 0.542 & 0.255 & \textbf{3.93} \\
PIRLNav~\cite{ramrakhya2023pirlnav} & Yes & \textbf{0.641} & 0.271 & - \\
\midrule
L3MVN (zero-shot)~\cite{yu2023l3mvn} & No & 0.504 & 0.231 & 4.43 \\
UIAP-OGN~\cite{bajpai2025uiapogn} & No & 0.523 & 0.290 & - \\
VLFM$^\dagger$~\cite{yokoyama2023vlfm} & No & 0.523 & 0.303 & 4.18 \\
RPV-V1 (ours) & No & 0.534 & 0.304 & 4.12 \\
RPV-V2 (ours) & No & \underline{0.539} & \textbf{0.307} & \underline{4.06} \\
\bottomrule
\end{tabular}
\end{table}

\subsection{ObjectNav Performance}
Table \ref{tab:objectnav_results} summarizes the navigation performance on the HM3D dataset. Among zero-shot approaches, our final proposed pipeline (RPV-V2) establishes highly competitive performance, achieving an SR of 0.539 and an SPL of 0.307. Compared to our primary zero-shot baseline, VLFM~\cite{yokoyama2023vlfm}, RPV-V2 demonstrates a $+3\%$ relative improvement in success rate and a $+1.3\%$ relative improvement in path efficiency (SPL). Furthermore, we outperform the recently proposed UIAP-OGN~\cite{bajpai2025uiapogn} framework in both metrics, displaying the robustness of our semantic room-probability projection. Notably, our pipeline achieves this zero-shot performance using only raw object-label strings as VLM inputs, thereby completely removing the need for explicit prompt engineering, text templates, or computationally expensive LLMs. This also highlights the distinct advantage of utilizing individual detected objects as spatial signals, rather than relying on noisy, image-holistic VLM embeddings.

While fully supervised methods like PIRLNav achieve a higher overall success rate, this advantage stems from extensive in-domain training via behavioral cloning on 77k human demonstrations collected in the HM3D training dataset split. Furthermore, because our pipeline is built directly on the 2D mapping framework of VLFM, we only support single-floor simulated environments and, as a result, cannot solve 14.6\% of HM3D validation episodes requiring multi-floor navigation. Despite this, RPV-V2 significantly outperforms both PIRLNav and the trained variant of L3MVN in SPL. By mathematically grounding frontier selection in geodesic distance fields rather than relying on black-box learned policies, our agent takes more direct paths to the target.

\subsection{Ablation of Signal Propagation}
\label{sec:signal_ablation}
To assess the influence of different signal‑propagation strategies on navigation, we compare against our RPV-V1 and RPV-V2 pipelines as described in Section~\ref{sec:metrics_and_baselines}.
As shown in Table~\ref{tab:objectnav_results}, changing the propagation method results in a slight absolute improvement of $+0.005$ points in SR, and $+0.003$ in SPL.

This comparison highlights the positives and negatives inherent in both propagation methods for embodied navigation. In RPV-V1, isotropic Gaussian signals bleed through walls, causing frontiers that are spatially disconnected from objects, yet inside the Gaussian radius, to accumulate low-magnitude responses. As frontier scoring uses a normalized sum, these weak but widespread signals suppress the influence of relevant object cues. However, as the signal is non-dependent on topology, detections made in ``unknown" space immediately influence navigation decisions, which is helpful to prevent bias towards close-proximity, high-object-density areas. By contrast, the FMM flood-fill ensures that signals wrap around obstacles and through doorways. As such, the signal cannot propagate into the ``unknown" but will more reliably affect frontiers spatially associated with the detected object. By enforcing this free-space topology, the resulting value map remains consistent with the reachable environment.

Despite the added mathematical rigor of FMM relative to simple isotropic stamping, our pipeline remains efficient, with an average inference time per step of 335.5~ms for RPV-V2. Although feasible for robotic deployment, resource‑constrained mobile platforms would benefit from offloading perception and VLM inference to an edge server.

\section{Conclusion}
We presented a training-free, open-vocabulary ObjectNav framework that maps semantic object knowledge directly to architectural room layouts in the CLIP embedding space, thereby exploiting estimated spatial co-occurrence likelihoods between objects. We introduce the concept of a Room Probability Vector (RPV) to provide spatial context for semantic relevance. By taking the dot product between RPVs, we computed the probabilistic overlap between the target and a detected object's room-probability distributions, projecting the score as a ``semantic signal" in a value map via flood-fill to respect geometric topology. This enables object‑centric frontier prioritization: searching regions containing objects that are more likely to co‑occur with the target. On the HM3D validation set, our flood-fill pipeline variant (RPV-V2) outperforms both a topologically unaware isotropic Gaussian propagation model (RPV‑V1) and the baseline comparison method in Success Rate (SR) and Success weighted by inverse Path Length (SPL).

Despite its effectiveness, existing limitations highlight promising directions for future work. First, the current RPV formulation does not explicitly account for entropy. Objects that appear across many room types (e.g., chairs or tables) naturally yield high-entropy room probability distributions, which can dilute the overall signal permeating the value map. Entropy-aware pruning or regularization could penalize these low-specificity signals, granting higher discriminatory power to functionally definitive objects (e.g., beds or bathtubs), increasing the dynamic range between good and bad frontiers for frontier selection. Second, the method for analytically estimating spatial co-occurrence does not strongly correlate with ground truth data. Investigating further models to reduce this gap would increase the reliability of signals. Representing the value map as a combination of isotropic Gaussians and flood-fill signals may also improve path efficiency.

Finally, the current navigation policy relies purely on semantic exploitation. In semantically sparse regions, such as hallways, the agent may over-commit to weak signals rather than seek novel vantage points or uncover more free space. Incorporating an information‑gain component could more effectively balance semantic guidance with topological exploration.

\subsection{Acknowledgements}
The authors acknowledge that AI-assisted tools (Gemini 3 Flash (Nano Banana 2) and ChatGPT) were used for limited editing of text and for generating some preliminary code and figure templates. All outputs were checked, corrected as needed, and integrated by the authors, who are solely responsible for the final content.

\bibliographystyle{IEEEtran}
\bibliography{references}

\end{document}